\documentclass{article}
\usepackage{acra}
\usepackage{graphicx}
\usepackage{booktabs}
\usepackage{url}
\usepackage{tabularx}
\usepackage{array}
\newcolumntype{Y}{>{\raggedright\arraybackslash}X}
\usepackage{amsmath}
\usepackage[font=small]{caption}

\usepackage{balance}
\usepackage[hidelinks]{hyperref}
\usepackage[T1]{fontenc}

\newcommand{\rev}[1]{#1}
\newcommand{\revj}[1]{#1}

\title{From Companion to Guardian: Designing an Emergency Support System Using a Quadruped Robot for Older Adults}

\author{%
  \textbf{Chunyang Meng\textsuperscript{1,2}\quad
  Eduardo B. Sandoval\textsuperscript{2}\quad
  Josh Andres\textsuperscript{3}}\\
  \textbf{Francisco Cruz\textsuperscript{1,4}\quad
  Mari Velonaki\textsuperscript{2}}\\[3pt]
  \textnormal{\textsuperscript{1} School of Computer Science and Engineering, UNSW Sydney, Australia}\\
  \textnormal{\textsuperscript{2} School of Art and Design, Creative Robotics Lab, UNSW Sydney, Australia}\\
  \textnormal{\textsuperscript{3} School of Systems and Computing, UNSW Canberra, Australia}\\
  \textnormal{\textsuperscript{4} Escuela de Ingenier\'ia, Universidad Central de Chile, Santiago, Chile}\\[3pt]
  \textnormal{Emails: \{chunyang.meng, e.sandoval, josh.andres\}@unsw.edu.au}\\
  \textnormal{\{f.cruz, mari.velonaki\}@unsw.edu.au}
}

\usepackage{etoolbox}

\date{}

\newcommand{\PaperCiteAs}{%
  Cite as: Chunyang Meng, Eduardo B. Sandoval,
  Josh Andres, Francisco Cruz, and Mari Velonaki.
  From Companion to Guardian: Designing an Emergency
  Support System Using a Quadruped Robot for Older Adults.
  \emph{Proceedings of the Australasian Conference
  on Robotics and Automation (ACRA)}, 2026.
}

\makeatletter
\pretocmd{\@maketitle}{%
  \begingroup
    \normalfont\small\bfseries
    \noindent\PaperCiteAs\par
  \endgroup
  \vspace{0.5em}
  \hrule
  \vspace{0.5em}
}{}{\PackageWarning{acra}{Failed to patch \string\@maketitle}}
\makeatother

\begin{document}
\maketitle

\begin{abstract}

As older adults increasingly live alone, emergency support becomes an important function for home companion robots. In Australia, demand for the care and support workforce is projected to nearly double between 2020--21 and 2049--50. Service robots such as robot dogs may support older adults by combining everyday companionship with assistance when human support is not immediately available. We present the design and implementation of a domestic companion-robot system on a Unitree Go1 quadruped platform that combines companionship with emergency awareness and response. The system uses a three-stage interaction model consisting of companion, danger-awareness, and emergency modes. \rev{A rule-based escalation policy} combines signs detected from visual and speech input to determine the current risk level and adjust how long the robot waits for the user to confirm safety before entering emergency mode. Additional detected signs can shorten this waiting time. Users can confirm safety, request immediate help, stop robot motion, or reset the system. \rev{We verified the implemented decision rules and escalation timing through 105 automated trials and 25 staged physical trials. Additional visual evidence shortened the waiting time, and the system entered emergency mode when the waiting time ended without confirmation.}

\end{abstract}

\section{Introduction}

As populations age, supporting older adults who live independently at home is becoming increasingly important. By 2046, between 3.4 and 4.0 million Australians are projected to live alone \cite{abs2024household}, while demand for the care and support workforce is expected to approximately double by 2049--50 \cite{nsc2021careworkforce}. Home-based assistive robots present an opportunity to complement periodic human care with support that remains available within the home.

Robotic care support becomes particularly important when a person is unable to request help. Falls affect approximately one in three people aged over 65 each year \cite{world_health_organization_falls_2021}, while other sudden events, such as stroke, cardiac arrest, or loss of consciousness, may similarly prevent a person from speaking, reaching a telephone, or activating an alarm. Timely assistance is critical, as a person who remains on the floor for an extended period after a fall faces increased risks of complications and death \cite{fleming_inability_2008}.

Existing wearable, ambient, and vision-based systems can detect falls and related signs of incapacitation with high accuracy under controlled conditions \cite{usmani_latest_2021,gorce_fall_2025}. Their effectiveness in the home depends on how they fit into everyday life. Older adults may forget or choose not to wear sensing devices because they are uncomfortable \cite{moore_older_2021}. Fixed cameras may be rejected because of concerns about privacy and continuous surveillance \cite{tham_barriers_2025}, and personal alarm systems still depend on users being willing and able to activate them \cite{heinbuchner_satisfaction_2010}. The challenge involves detecting a possible emergency, interpreting detected signs, and determining how a system should interact with the person before escalating to external assistance.

Companion robots could provide a platform for combining everyday interaction with safety support. Robots such as Paro and Aibo have demonstrated the potential to support companionship, social engagement, and emotional wellbeing \cite{gasteiger_friends_2021,friedman_hardware_2003}. Other systems, including the CompanionAble robot Hector and Hobbit, have combined companion robots with fall detection and caregiver notification \cite{schroeter_realization_2013,fischinger_hobbit_2016}, while mobile robots have also been paired with Wi-Fi sensing or wearable devices for collaborative fall response \cite{chen_collaborative_2024,liang_collaborative_2021}. These systems establish that a companion robot can carry a safety function. An important design question is: \revj{\textit{How should a domestic companion robot move from its everyday companion role into a safety and emergency-response role when signs of possible incapacity appear?}} This transition involves active confirmation, user control, escalation timing, and interpreting new signs as the situation develops.

\begin{figure*}[t]
    \centering
    \includegraphics[width=0.82\textwidth]{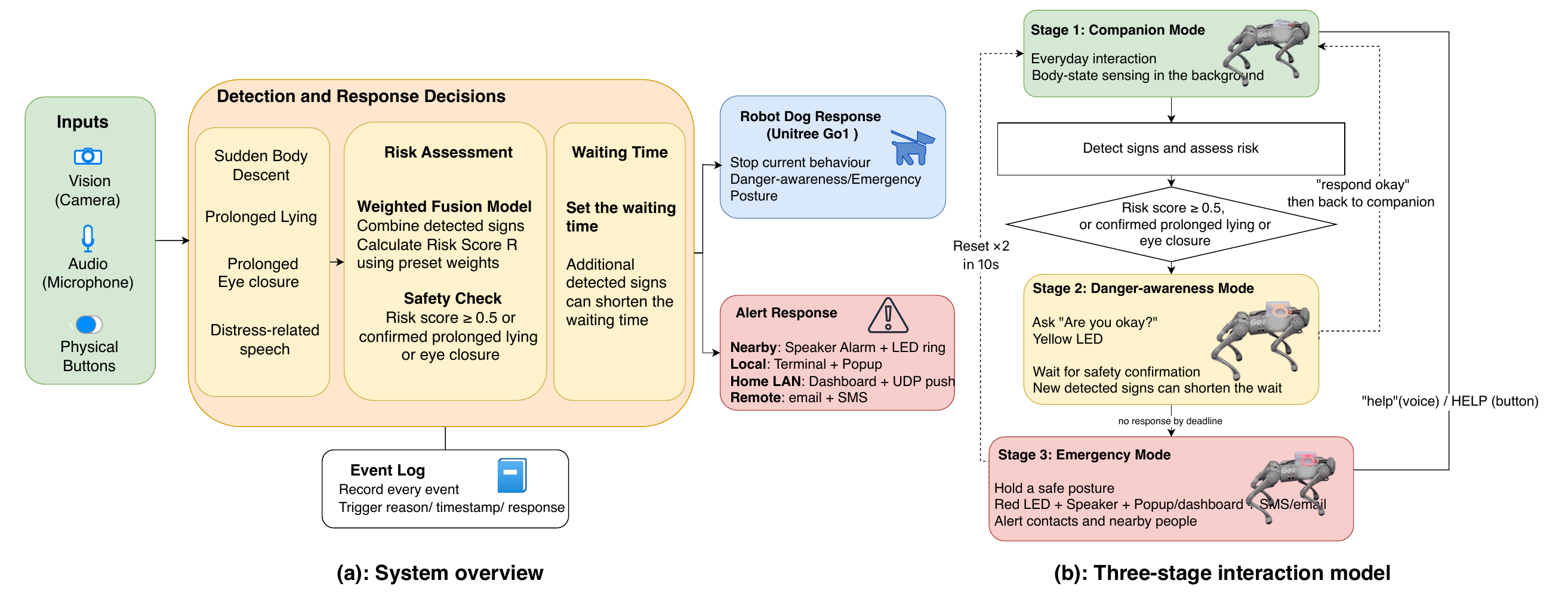}
    \caption{\rev{System overview.
    (a) System architecture: visual and speech input clients, physical buttons, the central emergency server, and the robot, LED, speaker, and alert outputs.
    (b) Three-stage interaction model showing the companion, danger-awareness, and emergency modes and the conditions for transitions between them.}}
    \label{fig:architecture}
\end{figure*}

In this paper, we present the design and implementation of a domestic companion-robot system on a Unitree Go1 quadruped platform. The system combines everyday companion interaction with emergency awareness and response. Figure~\ref{fig:architecture} presents the system overview and the three-stage interaction model. 
This work aims to make two contributions. 
\rev{First,} we introduce a three-stage interaction model consisting of companion, danger-awareness, and emergency modes. The model describes how the robot checks whether the user needs help and moves into emergency response. It also defines how users can confirm safety, request immediate help, stop robot motion, and reset the system.
\rev{Second,} we implement a system that combines signs detected from visual and speech input to initiate safety checks and adjust the waiting time before emergency response. These signs include sudden body descent, prolonged lying, prolonged eye closure, and distress-related speech. Additional detected signs can shorten the waiting time. If the user does not confirm safety before this time ends, the system enters emergency mode. \rev{We verify the implemented system} through 105 automated trials and 25 staged physical trials performed by one researcher.

\section{Related Work}

\subsection{Emergency Monitoring for Older Adults}

Research on emergency monitoring has produced three main families of detection technology: wearable sensors, ambient sensors, and vision-based systems \cite{usmani_latest_2021}. Many systems perform well in controlled or simulated settings \cite{gorce_fall_2025}. However, deployment in the home introduces challenges beyond sensing performance. Older adults may abandon wearable devices because they forget them or find them uncomfortable \cite{moore_older_2021}. Camera-based systems may meet resistance because they place continuous surveillance inside the home \cite{tham_barriers_2025}. False alarms can reduce trust in the system and the attention paid to subsequent alarms \cite{lewandowska_impact_2020}. Studies of personal emergency response systems add a further human factor. Many older adults who own an alarm pendant do not press it during a genuine emergency because they do not want to burden others \cite{heinbuchner_satisfaction_2010}. These findings show that effective home monitoring depends on both sensing and the design of the response.

Research in activity recognition has also examined the use of multiple sensing modalities to combine complementary evidence from different sources \cite{nweke_data_2019}. In emergency monitoring, multimodal interactive systems can combine observations such as body movement, posture, wearable sensing, and speech to provide multiple forms of evidence about a possible event. Recent robotic systems have also integrated \rev{wearable sensing}, visual verification, and dialogue within a fall-response architecture \cite{zoughalian_ros_2026}. This motivates an approach in which individual observations contribute to an evolving assessment of risk during the safety interaction.

\subsection{Companion Robots as Carriers of Safety Functions}

In the domain of companion robots, studies of the seal robot Paro report reduced loneliness and improved mood among older adults \cite{gasteiger_friends_2021}, while research on Aibo and similar robots shows that people can form lasting attachments to animal-like machines in their homes \cite{friedman_hardware_2003}. These robots can become entities with which people develop ongoing relationships \revj{\cite{bransky2026ai,andres2025scenario}, and these relationships can shape the conditions of their everyday worlds \cite{giaccardi_technology_2020,andres2026cosmogenic}}.

This relational role also creates opportunities to integrate safety functions into companion robots. The CompanionAble project combined its Hector robot with fall detection, emergency calls, and remote monitoring \cite{schroeter_realization_2013}. The Hobbit robot detects falls through changes in body posture and subsequently informs caregivers or emergency services \cite{fischinger_hobbit_2016}. Mobile companion robots have also been paired with Wi-Fi sensing \cite{chen_collaborative_2024} or wearable devices \cite{liang_collaborative_2021} to support collaborative fall detection and response. More recently, quadruped robots have also been explored as companions in older-adult care, including gesture-based interaction with older users \cite{meng_towards_2025}.

Together, these examples show how companion- and service-robot functions can be combined within a single robotic platform, allowing social and interactive roles to coexist with safety-related functions. Their response pipelines commonly include detection, user interaction, and notification. In our project, we focus on how evidence from the user's context accumulates after an initial sign of a potential emergency, how the robot checks whether assistance is required, and how the escalation timing changes as new evidence becomes available.

\subsection{Multimodal Interaction and Adaptive Response}

In emergency-response systems, spoken interaction can also provide information about the user's condition and allow the user to confirm or dismiss an alarm. Dialogue-based personal emergency response systems have used speech recognition to involve the user in deciding what assistance is required after an alarm is triggered \cite{hamill_development_2009}.

\rev{Work on mixed-initiative interaction recommends that a system use dialogue to resolve key uncertainties before acting, consider the cost of poorly timed actions, and allow users to intervene when it acts on their behalf \cite{horvitz_principles_1999}.} These principles inform our use of explicit user commands for confirmation, immediate assistance, stopping robot motion, and resetting the system.

Body-centric computing has explored the body as a site of interaction in contexts including performance, augmentation, and play \revj{\cite{mueller_next_2020,andres2023integrated,mueller2018experiencing}}. In an emergency, conventional input channels may become unavailable when a person cannot speak or reach a control. Posture, stillness, eye closure, and sudden body movement can then provide observable evidence of the person's condition. Our design combines these visual signs with distress-related speech to decide when to check on the user and how long to wait before entering emergency mode.

\section{\rev{System Design and Implementation}}
\label{sec:system-design}

In this system, evidence refers to detected signs that meet the detection and timing rules and are used for risk assessment. Camera images and microphone audio are the input data. Evidence does not confirm that an emergency has occurred.

The emergency server combines evidence, maintains the interaction mode, and updates the escalation deadline. This section describes the design principles, interaction model, and implementation \rev{on} the quadruped robot.

\subsection{Design Principles}

Four principles guide the transition from everyday companion behaviour to emergency response.

\textbf{Escalation}
A safety check gives the user time to confirm whether assistance is needed. \rev{As in dialogue-based personal emergency response systems \cite{hamill_development_2009},} the robot asks for a response before timed escalation. Additional evidence can change the urgency during this check and bring the deadline forward.

\textbf{User agency}
Users should be able to respond to a safety check and interrupt robot motion. Following mixed-initiative principles \cite{horvitz_principles_1999}, explicit commands provide control over confirmation, help, stopping, and reset. Leaving emergency mode requires confirmation so that a single accidental command does not clear the response.

\textbf{Multimodal redundancy}
A person's ability to respond may change during an emergency. Voice and physical buttons provide channels for deliberate input. Visual sensing provides evidence when the person cannot use these channels. The server combines the available observations during the safety interaction.

\textbf{Accountability}
Safety decisions should be traceable. The event log records the evidence received, the resulting policy level, and changes to the mode and deadline. These records connect the robot's response to the observations and commands that produced it and support the evaluation.

\subsection{Three-Stage Interaction Model}

Figure~\ref{fig:architecture}(b) shows the three interaction modes and the conditions for transitions between them. The system has three interaction modes: companion, danger-awareness, and emergency. Companion mode is the default operating state. When sufficient evidence of a possible emergency is detected, the system enters danger-awareness. If the user does not confirm safety before the active escalation deadline, the system transitions to emergency mode. Further details are provided below:

\textbf{Stage 1: Companion mode}
In its default state, the robot performs its normal interactive behaviours and can be controlled by the user, while safety sensing runs in the background. The visual client monitors sudden body descent, prolonged lying, and prolonged eye closure. These observations are temporally processed before they become evidence. The speech interface listens for user commands and distress-related phrases. The server combines the available evidence and determines whether a safety check should begin. A direct request for help enters emergency mode immediately.

\textbf{Stage 2: Danger-awareness mode}
The system enters danger-awareness when the policy level reaches HIGH, CRITICAL, or SEVERE through the weighted score or single-cue safety rule. The robot stops its current movement, adopts an attentive posture \rev{with its body slightly raised}, turns the LED yellow, and asks ``Are you okay?''. A local popup displays the current mode.

The server then waits for a user response, additional evidence, or the escalation deadline. An ``okay'' response or the OK button ends the episode and returns the system to companion mode. A ``help'' command or the HELP button enters emergency mode immediately. Additional evidence can update the policy level and shorten the deadline. Evidence expiry does not cancel the episode or extend its deadline. If the user does not confirm safety, the system enters emergency mode when the deadline is reached.

\textbf{Stage 3: Emergency mode}
Normal interaction is suspended. The robot stops its current movement, adopts a raised safety posture \rev{with its body lifted further into an alert stance (Fig.~\ref{fig:alert-responses}(a))}, turns the LED red, and produces an alarm and spoken warning. The server activates local and network alerts, with optional email and SMS notifications as well. In the current prototype, reset can be issued through the voice interface by saying ``reset'', or through the researcher keyboard interface using the corresponding reset key. The physical interface provides HELP and OK buttons but does not include a reset button. Leaving emergency mode requires two reset commands within 10~s.

The stop command is a motion-safety control. It immediately halts robot motion in any mode but does not change the interaction state or cancel the emergency alerts. In emergency mode, the LED, popup, network state, and alert pipeline therefore remain active, and an ongoing speaker warning is not silenced by the stop command.

\subsection{System Architecture}

Figure~\ref{fig:architecture}(a) shows the system architecture. The prototype consists of the Unitree Go1, an external webcam, an external speaker, an Arduino-controlled LED, physical HELP and OK buttons, and a laptop hosting the main software. The software uses separate input clients, a central emergency server, and a robot response module.

The visual client captures video with OpenCV and uses MediaPipe to estimate body and face landmarks. It applies the temporal qualification rules and sends evidence updates to the server. The speech interface uses an English Vosk model and recognises four explicit commands: ``help'', ``okay'', ``stop'', and ``reset''. It also recognises configured distress phrases including ``I fell'', ``I feel dizzy'', and ``I cannot get up'', together with equivalent shortened forms. An explicit ``help'' command enters emergency mode immediately. Distress phrases are converted into distress-speech evidence and participate in the weighted fusion policy. Distress-speech evidence has a weight of 0.50 and remains active for 10~s after it is received. Video and audio are processed locally on the host with offline models (MediaPipe and Vosk). Camera frames and audio are not stored or uploaded. Physical buttons send commands directly to the server. A keyboard interface supports development, testing, and recovery.

The server maintains active evidence, calculates the risk score, applies the single-cue rule, and manages the interaction mode and escalation deadline. It processes user commands according to the current mode. Pretrained models support perception, and the decision policy follows explicit rules.

The robot response module sends movement commands to the Go1. The server also coordinates the speaker, LED, and alert interfaces. Alert operations run independently so that a delay or failure in one channel does not block the local response. The event log records evidence, decisions, commands, and output actions with server timestamps.

\subsection{Temporal Evidence Qualification}

Table~\ref{tab:detection-rules} lists the evidence rules. The fall detector tracks a body centre calculated from shoulder and hip landmarks. Each positive check adds one to a confirmation counter. Each negative check subtracts one, with a minimum of zero. A count of three reports fall evidence. The 1.2~s window applies to body displacement.

The lying detector requires all four position and shape conditions to remain satisfied for 5~s. After activation, it allows a 1~s sensing dropout. A valid non-lying condition must persist for 1~s to clear the evidence. Eye closure uses the mean eye aspect ratio (EAR) from facial landmarks. Fall and distress-speech events remain available for fusion for 30 and 10~s, respectively.

\begin{table*}[t]
\centering
\small
\setlength{\tabcolsep}{5pt}
\begin{tabularx}{\textwidth}{@{}>{\raggedright\arraybackslash}p{2.3cm}Y>{\raggedright\arraybackslash}p{3.4cm}@{}}
\toprule
\textbf{Evidence} & \textbf{Activation rule} & \textbf{Timing} \\
\midrule
Fall
& Body-centre downward displacement $>0.30$ within 1.2~s; confirmation counter reaches 3
& Evidence active for 30~s \\
Prolonged lying
& All required: body-centre $y>0.72$; mean shoulder $y>0.62$; mean hip $y>0.68$; upper-body width/height $>2.0$
& Sustained for 5~s; 1~s dropout tolerance; 1~s clear duration \\
Prolonged eye closure
& Mean EAR $<0.18$
& Sustained for 5~s \\
Distress speech
& A configured distress phrase is recognised
& Evidence active for 10~s \\
\bottomrule
\end{tabularx}
\caption{Evidence rules. Positions use normalised image coordinates. Thresholds were set empirically for the laboratory setup.}
\label{tab:detection-rules}
\end{table*}

\subsection{\rev{Multimodal Evidence Fusion and Policy Levels}}

The emergency server maintains four safety evidence variables: fall, prolonged lying, prolonged eye closure, and distress-related speech. It tracks the current risk score, policy level, interaction mode, and escalation deadline. The score represents the evidence that is currently active. The policy level determines the no-response window. The mode controls the robot behaviour and alert outputs. The deadline specifies when the current danger-awareness episode will enter emergency mode if the user does not confirm safety.

The active evidence is combined using a weighted risk score:
\begin{equation}
\begin{aligned}
R(t)=\min\{1,\;&0.55F(t)+0.35L(t)\\
              &+0.25E(t)+0.50V(t)\},
\end{aligned}
\label{eq:risk}
\end{equation}
where $F(t)$ represents fall evidence, $L(t)$ represents qualified lying evidence, $E(t)$ represents qualified eye-closure evidence, and $V(t)$ represents distress-related speech evidence. Each variable is binary and indicates whether the corresponding evidence is currently active.

The weights are heuristic parameters selected to encode the intended escalation structure. Fall evidence and distress-related speech are each sufficient to reach the HIGH threshold. Combinations of active evidence can reach CRITICAL or SEVERE. Lying combined with eye closure produces a score of 0.60 and remains HIGH. Because this combination is also consistent with sleep, it triggers a safety check but does not shorten the no-response window. The score-based level is
\begin{equation}
q_{\mathrm{score}}(t)=
\begin{cases}
\text{SEVERE}, & R(t)\geq0.90,\\
\text{CRITICAL}, & 0.70\leq R(t)<0.90,\\
\text{HIGH}, & 0.50\leq R(t)<0.70,\\
\text{MONITOR}, & 0<R(t)<0.50,\\
\text{CLEAR}, & R(t)=0.
\end{cases}
\label{eq:risk-level}
\end{equation}

The policy level $q(t)$ also applies a single-cue safety rule. Qualified lying or eye closure can independently set the policy level to HIGH when the score is below 0.50. The score remains unchanged. Eye closure alone therefore produces a score of 0.25, a score-based level of MONITOR, and a policy level of HIGH. Lying alone produces a score of 0.35 and also sets HIGH. In other cases, the policy level follows the score-based level.

For example, fall evidence produces a score of 0.55 and a HIGH policy level. If qualified eye closure becomes active while fall evidence remains available, the score becomes 0.80 and the level becomes CRITICAL. Fall combined with qualified lying produces a score of 0.90 and a SEVERE level.

When the user responds ``okay'', the system ends danger-awareness and clears stored fall and distress-speech evidence. Any active lying or eye-closure condition is acknowledged and suppressed until it clears or its input client disconnects. This prevents the same continuous condition from immediately starting another episode. A new fall or distress-speech event can initiate a new episode.

\subsection{Adaptive Escalation Policy}

Each escalation level has a configurable no-response window:
\begin{equation}
T(q(t))=
\begin{cases}
20~\text{s}, & q(t)=\text{SEVERE},\\
40~\text{s}, & q(t)=\text{CRITICAL},\\
60~\text{s}, & q(t)=\text{HIGH}.
\end{cases}
\label{eq:adaptive-window}
\end{equation}

When danger-awareness begins at time $t_0$, the server sets the deadline to $t_0+T(q(t_0))$. Further evidence processed at time $t$ updates the deadline according to
\begin{equation}
d_{\mathrm{new}}=\min\bigl(d_{\mathrm{current}},\;t+T(q(t))\bigr),
\label{eq:deadline}
\end{equation}
where $d_{\mathrm{current}}$ is the existing deadline and $T(q(t))$ is the window for the current policy level. This update applies when the policy level is HIGH, CRITICAL, or SEVERE. CLEAR and MONITOR leave the existing deadline in place.

The minimum operation allows new evidence to shorten the deadline and prevents it from being extended. If evidence later expires or clears, the server retains the deadline already established for the episode. The current score and policy level continue to reflect the active evidence.

For example, fall evidence sets a HIGH-level deadline 60~s after danger-awareness entry. The evidence expires after 30~s. With no other evidence, the score becomes zero and the current policy level becomes CLEAR. The system remains in danger-awareness and enters emergency mode at the original deadline if the user does not confirm safety. A help request enters emergency mode immediately, and an ``okay'' response ends the episode. A stop command leaves the safety state active. These commands are handled separately from the weighted score.

\subsection{Robot Behaviour and Alert Responses}

\begin{figure*}[t]
    \centering
    \includegraphics[width=0.8\textwidth]{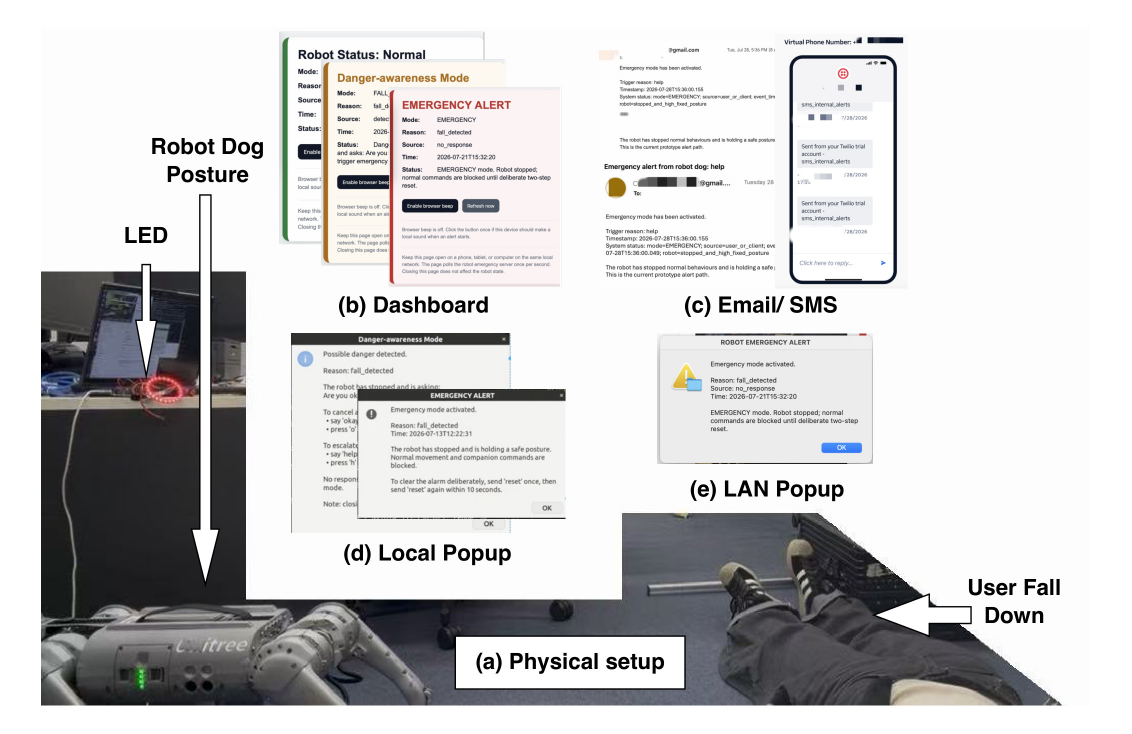}
    \caption{Prototype setup and alert interfaces.
    (a) Robot posture and LED during a staged fall trial.
    (b) Dashboard.
    (c) Email and SMS notifications.
    (d) Local popup.
    (e) Popup on another device connected to the same local
    area network (LAN).}
    \label{fig:alert-responses}
\end{figure*}

Figure~\ref{fig:alert-responses} shows the physical implementation and alert interfaces.

The host presents a local popup and terminal notification. A home-network dashboard displays the current mode, trigger reason, and timestamp. State changes are broadcast through UDP to listener devices on the same network. Optional email and SMS interfaces notify configured contacts. These remote channels are separate from the state transition logic. Alert operations run independently, and a delay or failure in one output channel does not block the robot's local response. Failures are recorded in the event log.

The log records incoming evidence, current scores and policy levels, user commands, mode transitions, deadline changes, and alert actions. It also records the score and policy level used to set the active deadline. These timestamps provide a record of each decision and the measurements used in Section~\ref{sec:evaluation}.

\section{Evaluation}
\label{sec:evaluation}

\rev{We verified the system at two levels.} Automated trials tested evidence fusion and escalation using predefined inputs. Staged physical trials tested whether the visual perception pipeline could produce the evidence needed for the intended transitions. All timing measurements came from timestamps recorded by the emergency server.

\subsection{Evaluation Procedure}

The automated evaluation injected predefined evidence events directly into the final emergency server. This allowed the decision policy to be tested without changes in camera position, body movement, or speech recognition. The tests covered single-cue activation, multimodal combinations, deadline updates, and the rule that a shortened deadline cannot be extended.

There were 21 configurations with five repetitions each: nine policy configurations (45 trials), four deadline configurations (20 trials), and eight fixed/adaptive configurations across four sequences (40 trials). The server used its full 60, 40, and 20~s windows with a time scale of 1.0. These 105 trials did not include dedicated tests of help, okay, stop, or reset. These commands were tested in video-recorded demonstrations and worked as expected, but were not included in the logged trial set.

The physical evaluation used the visual perception and emergency-response pipeline in the laboratory. \rev{The webcam was placed in front of the participant at a height of approximately 0.85~m and a distance of approximately 1.2~m, facing the participant. Eye-closure detection requires the face to be visible to the camera, which may not hold when a person lies on the floor facing away from or far from the camera.} One researcher performed five conditions: fall only, eye closure only, lying only, fall followed by eye closure, and fall followed by lying. Each condition was repeated five times, giving 25 trials. In the fall-only condition, the researcher performed a rapid controlled descent and then remained seated with eyes open. These trials produced fall evidence without subsequent lying or eye-closure evidence.

The measurements recorded evidence receipt, policy-level changes, deadline updates, and entry into emergency mode. Physical timing began when the server received the first relevant evidence and ended at the emergency transition. Initial perception latency and remote message delivery were outside this interval. \rev{The trials assessed the implemented transitions and timing. They did not measure perception accuracy across users or environments.}

The command test and physical trial demonstrations are available in the
\href{https://drive.google.com/drive/folders/1mDaYWwSdVo7g5eiz9CsXaivFvB-nCRUE?usp=sharing}{\underline{supplementary video}}.

\subsection{\rev{Automated Policy Verification}}

All 105 automated trials passed their expected assertions. The nine policy configurations covered the four individual cues and five combinations: fall with eyes, fall with lying, fall with distress speech, lying with eyes, and voice with lying.

The policy tests examined scores, effective policy levels, and response windows. Fall evidence alone produced a score of 0.55 and selected HIGH with a 60~s window. Fall with eye closure produced 0.80 and selected CRITICAL with 40~s. Fall with lying produced 0.90 and selected SEVERE with 20~s. Qualified eye closure and lying independently selected HIGH through the single-cue rule. Their scores remained 0.25 and 0.35, respectively.

The fixed/adaptive tests used four sequential evidence conditions while the server was in danger-awareness. Table~\ref{tab:auto-results} reports the mean scheduled deadlines from mode entry across five trials per configuration. The fixed policy retained a 60~s deadline. The adaptive policy updated the deadline using the current level. These tests checked scheduling and ended before entry into emergency mode.

\begin{table}[t]
\centering
\small
\setlength{\tabcolsep}{4pt}
\begin{tabularx}{\columnwidth}{@{}Yrr@{}}
\toprule
\textbf{Sequence} & \textbf{Fixed (s)} & \textbf{Adaptive (s)} \\
\midrule
Eyes $\rightarrow$ lying & 60.00 & 60.00 \\
Fall $\rightarrow$ eyes & 60.00 & 42.08 \\
Fall $\rightarrow$ lying & 60.00 & 23.08 \\
Fall $\rightarrow$ voice & 60.00 & 21.58 \\
\bottomrule
\end{tabularx}
\caption{Mean scheduled deadlines from danger-awareness entry; five trials per configuration.}
\label{tab:auto-results}
\end{table}

Eyes followed by lying retained the HIGH level and the 60~s deadline. In the fall-to-eyes sequence, the additional evidence arrived approximately 2.08~s after danger-awareness entry. It selected the 40~s CRITICAL window, giving a mean deadline of 42.08~s. Lying arrived approximately 3.08~s after entry and selected the 20~s SEVERE window, giving 23.08~s. Distress speech arrived approximately 1.58~s after entry and also selected SEVERE, giving 21.58~s. The SDs of these three scheduled deadlines were 0.0008, 0.0011, and 0.0003~s, respectively.

The four deadline configurations tested the three fall sequences and the rule that an active deadline cannot be extended. Timer error was the actual emergency transition time minus the scheduled deadline. Mean errors were 0.0008~s (SD 0.0001~s) for fall then eyes, 0.0007~s (SD $<0.0001$~s) for fall then lying, and 0.0008~s (SD 0.0001~s) for fall then voice. Evidence reductions did not extend a shortened deadline. These results confirmed the rules in Section~\ref{sec:system-design} and the execution of the tested deadlines.

\subsection{\rev{Physical Scenario Verification}}

The physical evaluation examined whether the escalation policy operated with evidence from the implemented vision pipeline. Table~\ref{tab:physical-results} summarises the five conditions, with five repetitions per condition.

\begin{table}[t]
\centering
\small
\setlength{\tabcolsep}{4pt}
\begin{tabularx}{\columnwidth}{@{}Yrr@{}}
\toprule
\textbf{Condition} & \textbf{Mean (s)} & \textbf{SD (s)} \\
\midrule
Fall only & 60.00 & $<0.01$ \\
Eye closure only & 60.00 & $<0.01$ \\
Lying only & 60.00 & $<0.01$ \\
Fall $\rightarrow$ eye closure & 45.78 & 0.29 \\
Fall $\rightarrow$ lying & 26.86 & 0.49 \\
\bottomrule
\end{tabularx}
\caption{Time from initial evidence receipt to emergency mode; five trials per condition.}
\label{tab:physical-results}
\end{table}

The three single-cue conditions selected HIGH and reached emergency mode after approximately 60~s. The fall-only deadline remained active after fall evidence expired. Qualified eye closure and lying activated HIGH through the single-cue rule.

After fall evidence, eye closure arrived at $5.78\pm0.29$~s, raised the score to 0.80, and selected CRITICAL with a 40~s window. Emergency mode began at $45.78\pm0.29$~s. Lying arrived at $6.86\pm0.49$~s, raised the score to 0.90, and selected SEVERE with a 20~s window. Emergency mode began at $26.86\pm0.49$~s. Both intervals are measured from the initial fall evidence.

\section{Discussion}

\subsection{Evidence-Based Escalation}

The staged tests demonstrate how evidence updates the policy level and active deadline. Retaining the deadline allows escalation after evidence expires. Additional evidence can shorten the waiting period. Physical response times also depend on when qualified evidence reaches the server. The reported timer errors describe server execution. Full response timing requires measurements from physical event onset to remote message receipt.

\subsection{\rev{Expected False Alarms in Everyday Activities}}

\rev{False alarms can reduce trust in monitoring systems \cite{lewandowska_impact_2020}, so the behaviour of the current thresholds during everyday activities needs to be considered. The fall cue requires a large downward displacement within 1.2~s. In the fall-only condition, the researcher performed a deliberately rapid descent that was more forceful than ordinary sitting. Older adults typically sit down more slowly and in a more controlled way, so everyday sitting is less likely to meet this threshold, although this was not tested. The single-cue rule raises a more likely source of false alarms. Dozing in a chair can produce eye-closure evidence, and lying on a sofa can produce lying evidence. Each of these starts a 60~s safety check. Lying combined with eye closure, which is consistent with sleep, remains HIGH and does not shorten the window. In these cases, the user is expected to end the check by saying ``okay'' or pressing the OK button. The system then suppresses the acknowledged lying or eye-closure condition until it clears, so a continuous rest produces one check rather than repeated checks. However, a person who is asleep may not hear or answer the check, and the system would then enter emergency mode after 60~s. This remains a false-alarm risk. Contextual information, such as whether the person is on a bed or sofa or the time of day, and a repeated spoken prompt before emergency mode could reduce it. Negative trials with everyday activities that should not escalate, such as sitting down, resting with eyes closed, and lying on a sofa, are needed to measure this behaviour.}

\subsection{Interaction and User Control}

Danger-awareness provides time for user confirmation before emergency response. Voice and buttons support direct commands, and visual sensing provides evidence when the user cannot respond. Acknowledged persistent cues remain suppressed until they clear. Evaluation with older adults and caregivers is needed to assess perceived control and acceptable escalation timing.

\subsection{\rev{Role of the Quadruped Platform}}
\rev{In the current prototype, sensing, processing and alerts are external to the Go1. The robot body contributes in two ways. First, the safety check is delivered by a companion the user already interacts with, which may make it more acceptable than a request from a monitoring device. Second, the robot's posture change makes the mode transition visible. In the planned version with an onboard camera and navigation, mobility would allow the robot to approach the user and reposition the camera when a fixed view is occluded. These potential benefits have not yet been evaluated.}

\subsection{Limitations and Future Work}

The weights, thresholds, and windows were set manually. The score is not a calibrated emergency probability, and related cues contribute separately. Future work should examine calibration, cue dependence, and personalised timing. 

Physical trials involved one researcher and a fixed camera. Researcher familiarity and staged actions may have influenced the observations. Perception accuracy across users, lighting, occlusion, and homes was not measured. The help, okay, stop, and reset commands were demonstrated on the final configuration, but early interaction and protection checks were not repeated as logged trials. Dedicated tests of help, okay, stop, reset, and duplicate-trigger handling remain necessary. The experiments do not establish user trust, perceived control, or everyday reliability. Remote message receipt was not measured.

The prototype does not use autonomous navigation for emergency response, and the camera is external to the robot. Future work will mount the camera on the Go1 and should examine navigation, additional sensing, and evaluation with older adults and caregivers, including privacy, false alarms, and user control.

\section{Conclusion}

We presented a Unitree Go1 companion robot with companion, danger-awareness, and emergency modes. Multimodal evidence sets an adaptive escalation deadline \rev{that can only shorten}, and users can confirm safety or request help. All 105 automated trials passed their assertions\rev{, and the 25 staged physical trials followed the intended} HIGH, CRITICAL, and SEVERE escalation paths. \rev{These results verify the implemented policy. Future work will evaluate detection accuracy and false alarms through negative trials in everyday settings, and examine calibration, navigation, and evaluation with older adults and caregivers.}

\balance

\bibliographystyle{named}
\bibliography{emergency}
\end{document}